%% file: acl_latex.tex
\pdfoutput=1
\documentclass[11pt]{article}
\usepackage{acl}

\usepackage{times}
\usepackage{latexsym}

\usepackage{graphicx}
\usepackage{amssymb}
\usepackage{booktabs}
\usepackage{multirow}
\usepackage{multicol}
\usepackage{hyperref}

\usepackage{etoolbox}
\usepackage{tikz}

\usepackage{cleveref}
\crefname{section}{§}{§§}
\Crefname{section}{§}{§§}
\newcommand\pcref[1]{(\Cref{#1})}
\newcommand\cm{\checkmark}

\usepackage{listings}
\usepackage{framed}

\usepackage{xcolor,soul}

\newcommand{\myname}{DISAPERE}

\newcommand{\coarse}{\textsc{review-action} }
\newcommand{\fine}{\textsc{fine-review-action} }
\newcommand{\aspect}{\textsc{aspect} }
\newcommand{\polarity}{\textsc{polarity} }

\newcommand{\rebfine}{\textsc{rebuttal-action} }
\newcommand{\rebcoarse}{\textsc{rebuttal-stance} }

\usepackage{tikz}
\usetikzlibrary{tikzmark}

\usepackage{makecell}
\usepackage[T1]{fontenc}
\usepackage[utf8]{inputenc}

\usepackage{microtype}

\title{DISAPERE: A Dataset for Discourse Structure in Peer Review Discussions}

\author{Neha Nayak Kennard \qquad Tim O'Gorman  \qquad Rajarshi Das \\
\textbf{Akshay Sharma} \qquad \textbf{Chhandak Bagchi} \qquad  \textbf{Matthew Clinton} \qquad \\
\textbf{ Pranay Kumar Yelugam} \qquad \textbf{Hamed Zamani} \qquad \textbf{Andrew McCallum} \\
University of Massachusetts Amherst\\
\texttt{\{kennard, togorman, rajarshi, akshaysharma, cbagchi, }\\
\texttt{mfclinton, pyelugam, zamani, mccallum\}@cs.umass.edu}}

\begin{document}
\maketitle

\input{00_abstract}
\input{01_intro}
\input{02_related}
\input{03_dataset}
\input{04_analysis}
\input{05_application}
\input{06_experiments}

\input{07_conclusion}

\input{10_acknowledgements}

\bibliography{anthology,custom}
\bibliographystyle{acl_natbib}

\appendix

\input{08_appendices}

\end{document}

%% file: 00_abstract.tex
\begin{abstract}

At the foundation of scientific evaluation is the labor-intensive process of peer review. This critical task requires participants to consume vast amounts of highly technical text. Prior work has annotated different aspects of review argumentation, but discourse relations between reviews and rebuttals have yet to be examined.

We present DISAPERE, a labeled dataset of 20k sentences contained in 506 review-rebuttal pairs in English, annotated by experts. DISAPERE synthesizes label sets from prior work and extends them to include fine-grained annotation of the rebuttal sentences, characterizing their context in the review and the authors' stance towards review arguments. Further, we annotate \textit{every} review and rebuttal sentence.

We show that discourse cues from rebuttals can shed light on the quality and interpretation of reviews. Further, an understanding of the argumentative strategies employed by the reviewers and authors provides useful signal for area chairs and other decision makers.

\end{abstract}

%% file: 01_intro.tex
\section{Introduction}

\begin{figure*}[ht]
    \centering
    \includegraphics[width=\textwidth]{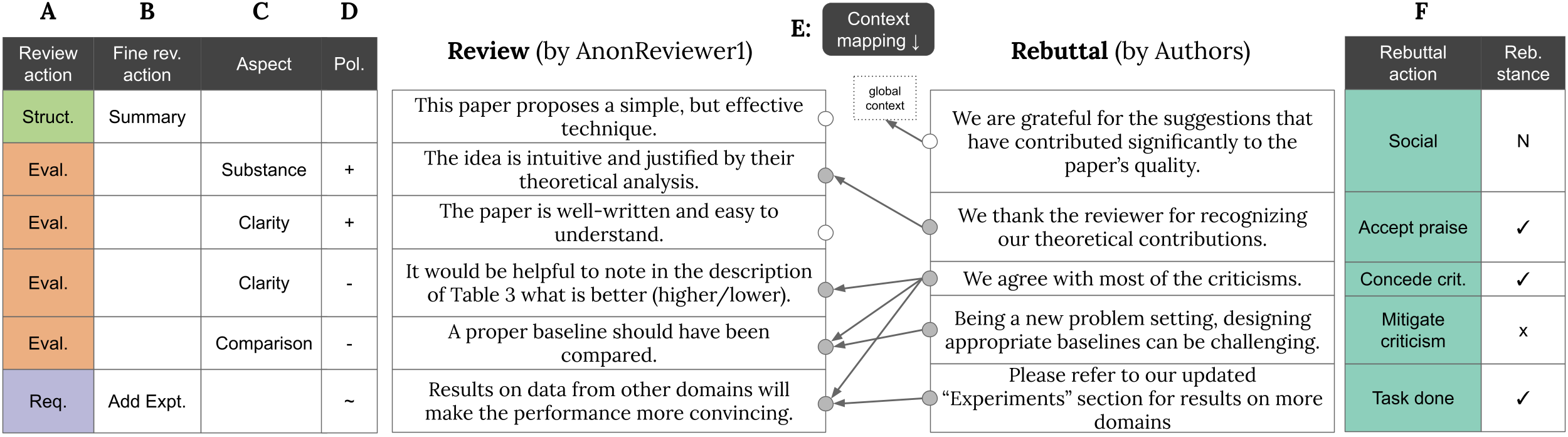}
    \caption{A depiction of our annotation scheme on a minimal, fictional review-rebuttal pair. A: \coarse, including Structuring, Request, Evaluation; B: \fine, fine-grained categorization of Structuring and Request sentences; C: \aspect, indicating the qualities of the manuscript being commented upon D: \polarity indicating whether these comments are positive or negative in nature.
    E: Each sentence in the rebuttal is mapped to zero or more sentences in the review, which constitute its context. This is a many-to-many relation. F: The sentences in the rebuttal are labeled with domain-specific discourse acts (\rebfine); each discourse act may be categorized according to whether it concurs with (\checkmark) or disputes ($\times$) the premise of the context it is responding to.}
    \label{fig:mainfig}
\end{figure*}

Peer review performs the essential role of quality control in the dissemination of scientific knowledge. The recent rapid increase in academic output places an immense burden on decision makers such as area chairs and editors, as their decisions must take into account not only extensive manuscripts, but enormous additional amounts of technical text including reviews, rebuttals, and other discussions.

One long term goal of research in peer review is to support decision makers in managing their workload by providing tools to help them efficiently absorb the discussions they must read. While machine learning should not be used to produce condensed accounts of the peer review text due to the risk of amplifying biases \citep{zhao-etal-2017-men}, ML tools could nevertheless help manage information overload by identifying patterns in the data, such as argumentative strategies, goals, and intentions. 

Any such research requires an extensive labeled dataset. While the OpenReview platform \citep{soergel2013open} has made it easy to obtain unlabeled public peer review text, labeling this data for supervised NLP requires highly qualified annotators. Correct interpretation of the discourse structure of the text requires an understanding of the technical content, precluding the use of standard crowdsourcing techniques. Prior work on discourse in peer review has focused this qualified labor force on labeling arguments extracted from the text, which enables the complete annotation of more examples, at the expense of research on non-argumentative behaviors in peer review. While there has been extensive research and deep analysis of different aspects of peer review, the taxonomies used to describe review argumentation are disparate and not directly compatible. Finally, there has been limited research into understanding the discourse relations between rebuttals and reviews \citep{cheng-etal-2020-ape, bao-etal-2021-argument}, and none so far into the discourse structure of rebuttals.

This paper presents \textbf{DISAPERE} (\textbf{DI}scourse \textbf{S}tructure in \textbf{A}cademic \textbf{PE}er \textbf{RE}view), a dataset focusing on the interaction between reviewer and author\footnote{The dataset, along with its accompanying code and documentation, is available at \url{http://www.github.com/nnkennard/DISAPERE/}.}. We give reviews and rebuttals equal importance, and emphasize the relations between them. To enable the study of behaviors beyond the core arguments, we also annotate every sentence of both the review and rebuttal, and provide fine-grained labels for non-argumentative types. We annotate at the sentence level not only for completeness but also to avoid the propagation of errors from argument detection. We annotate four properties (\textsc{review-action}, \textsc{fine-review-action}, \textsc{aspect}, \textsc{polarity}) of each review sentence, where the set of properties and their values were developed by synthesizing taxonomies from prior work. We also annotate each sentence of a rebuttal with a fine-grained label indicating the author's intentions and commitment, and a link to the set of review sentences that form its context.  \Cref{fig:mainfig} shows the DISAPERE annotation scheme on a minimal, fictional example review-rebuttal pair.

DISAPERE is intended as a comprehensive and high-quality test collection, along with training data to fine-tune models.  Our annotations are carried out by graduate students in computer science who have undergone training and calibration, amounting to over 850 person-hours of annotation work. Much of the test data is double-annotated, and we report inter-annotator agreement on all aspects of the annotation. We describe the performance of state of the art models on the tasks of predicting labels and contexts, showing that interesting ambiguities in the data provide the NLP community with research challenges.  We also show an example that demonstrates how decision makers could use models like these to understand trends and inform policies for future conferences  \pcref{sec:applications}.

The contributions of this paper are as follows:
(1) a new labeled training dataset of 506 review-rebuttal pairs (over 20k sentences) of peer review discussion text in English, where review sentences are annotated with four properties, and rebuttal sentences are annotated with context and labels from a novel scheme to describe discourse structure;
(2) a taxonomy of discourse labels synthesizing prior work on discourse in peer review and extending it to add useful subcategories;
(3) a summary of the performance of baseline models on the dataset \pcref{sec:baselines};
(4) examples of analyses on the dataset that could benefit peer review  decision makers \pcref{sec:analysis,sec:applications}, and
(5) extensive annotation guidelines and software to support future labeling efforts. 

%% file: 02_related.tex
\section{Related work}

The design of this dataset draws upon extensive, but disparate prior work on this topic. Many works, some addressed below, have taken advantage of the availability of review text hosted on OpenReview.

\paragraph{Argument-level review labeling} Prior work has developed label sets that address different phenomena. \citet{hua-etal-2019-argument-mining} introduced the study of discourse structure in peer review by annotating argumentative propositions in the AMPERE dataset with a set of labels tailored to the peer review domain (\textsc{evaluation}, \textsc{request}, \textsc{fact}, \textsc{reference}, and \textsc{quote}). Similarly, \citet{fromm2020argument}'s AMSR dataset frames the problem as an argumentation process, in which the stance of each argument towards the paper's acceptance or rejection is of paramount importance. Both view peer review as argumentation, using argument mining techniques to highlight spans of interest.

While its goal is not to examine discourse structure per se, \citet{yuan2021automate} uses polarity labels to indicate each argument's support or attack of the authors' bid for acceptance. Besides polarity, these examples follow \citet{Chakraborty-2020} by annotating each argument with the \textit{aspect} of the paper it comments on.\footnote{Aspects are based on the \href{https://web.archive.org/web/20211021081332/https://acl2018.org/downloads/acl_2018_review_form.html}{ACL 2018 rubric.}} In contrast to \citet{yuan2021automate}, we do not attempt or recommend generating peer review text, instead focusing on \textit{analyzing} human-generated text in peer review.

\paragraph{Review-rebuttal interactions}

We also expand on work by \citet{cheng-etal-2020-ape}, who first annotated discourse relations between sentences in reviews and rebuttals. While \citet{cheng-etal-2020-ape, cheng-etal-2021-argument} present new deep learning architectures, in this paper we focus on the creation and comprehensive annotation of a new dataset, illustrated with results from some less specialized baseline models.

Other research into rebuttals includes \citet{gao-etal-2019-rebuttal}. Besides their main finding that reviewers rarely change their rating in response to rebuttals, they find that more specific, convincing and explicit responses are more likely to elicit a score change.
Observations from this paper are formalized into rebuttal action labels in DISAPERE.

\paragraph{Comparison of datasets} In \myname\, we attempted to unify these schemas to form a single hierarchical schema for review discourse structure. We then expanded this hierarchical schema to introduce fine-grained classes for implicit and explicit requests made by the reviewers. The details of the correspondence between \myname\ labels and those from prior work are summarized in \Cref{sec:rationale}. In contrast to prior work, \myname\ labels discourse phenomena at the sentence level rather than the argument level. This enables more thorough coverage of the text while avoiding the propagation of errors from machine learning models earlier in the annotation pipeline. While using manually defined discourse units (above or below the sentence level) may more precisely capture some  discourse information, a separate pass of discourse segmentation can hinder the use of discourse datasets, as achieving consistent and replicable annotation of argument units is known to be highly challenging~\citep{trautmann2020fine}, and also because few works actually tackle unit segmentation~\citep{ajjour-etal-2017-unit}.

\input{tables/06a_related_table.tex}

%% file: tables/06a_related_table.tex
\begin{table}[ht]
\resizebox{\columnwidth}{!}{
\begin{tabular}{@{}lllllll@{}}
\toprule
	& Dataset & \rotatebox{90}{AMPERE} & \rotatebox{90}{AMSR} & \rotatebox{90}{ASAP-Review} & \rotatebox{90}{APE} & \rotatebox{90}{\textbf{\myname}} \\ \midrule
\multicolumn{2}{l}{\# examples} & 400 & 77 & 1k & 4.7k & 506 \\
\multicolumn{2}{l}{\# labels} & 10k & 1.4k & 5.7k & 130k & 46k \\ \midrule
\multirow{6}{*}{\rotatebox{90}{Review}}  &
Arg. stmts. & \cm & \cm & & \cm & \cm \\
 & Arg. types & \cm & & & & \cm \\ 
 & Polarity & & \cm & \cm & & \cm \\
 & Aspect & & & \cm & & \cm \\
 & Non-arg. & & & & & \cm \\
 & All sents. & & & & & \cm \\ \midrule
\multirow{6}{*}{\rotatebox{90}{Rebuttal}} &
Included? & & & & \cm & \cm \\
 & Arg. stmts. & & & & \cm & \cm \\
 & Context & & & & \cm & \cm \\
 & Arg. types & & & & & \cm \\
 & Non-arg. & & & & & \cm \\
 & All sents. & & & & & \cm \\ \bottomrule
\end{tabular}
}
    \caption{\label{tab:comparison}Comparison between our dataset and prior work: AMPERE \citep{hua-etal-2019-argument-mining}, AMSR \citep{fromm2020argument}, ASAP-Review \citep{yuan2021automate}, APE \citep{cheng-etal-2020-ape}.
    \textit{Arg.stmts.}: Are argumentative statements highlighted?; \textit{Arg. types}: Are subtypes of argumentative statements labeled?; \textit{Non-arg}: Are non-argumentative statements labeled?; \textit{All sents.}: Are labels provided for all sentences?; \textit{Context}: Are rebuttal texts' contexts in the review  provided? \myname\ is the only work to annotate every sentence in the review and rebuttal, and the only work that applies discourse labels to the author's actions in the rebuttal.} 
    
\end{table}

%% file: 03_dataset.tex
\section{Dataset}
\label{sec:dataset}

Each example in \myname\ consists of a pair of texts: a review and a rebuttal. Labels for reviews and rebuttal sentences are described below. Review sentence labels are summarized in \Cref{tab:reviewlabels}, and rebuttal sentence labels in \Cref{tab:rebuttallabels}.

\subsection{Review sentence labels}

\input{tables/02a_description_table}

\subsubsection{Review actions}

\coarse annotations characterize a sentence's intended function in the review. Annotators label each sentence with one of six coarse-grained sentence types including  \textit{evaluative} and \textit{fact} sentences,  \textit{request} sentences (including questions, which are requests for information), as well as non-argument types: \textit{social}, and \textit{structuring} for organization of the text.  

\subsubsection{Fine-grained review actions}
We also extend two of these review actions with subtypes: \textit{structuring} sentences include headers, quotations, or summarization sentences, and \textit{request} sentences are subdivided by the nature of the request, distinguishing between  clarification of factual information, requests for new experiments, requests for an explanation (e.g. of motivations or claims), requests for edits, and identification of minor typos.
 
\subsubsection{Aspect and polarity} \textsc{aspect} annotations follow the ACL review form \citep{ Chakraborty-2020, yuan2021automate}. These distinguish \textit{clarity}, \textit{originality}, \textit{soundness/correctness}, \textit{replicability}, \textit{substance}, \textit{impact/motivation}, and \textit{meaningful comparison}. Following  \citet{yuan2021automate}, arguments with an \textsc{aspect} are also annotated for \textsc{polarity}. We label \textit{positive} and \textit{negative} polarities. \textsc{aspect} and \textsc{polarity} are applied to sentences whose \textsc{review-action} value is \textit{evaluative} or \textit{request}.

\subsection{Rebuttal sentence labels}

\input{tables/02aa_rebuttal_description}
We annotate two properties of each rebuttal sentence: a \textsc{rebuttal-action} label characterizing its intent, and its \textsc{context} in the review in the form of a subset of review sentences.

% Rebuttal sentences are annotated with labels at two levels of a taxonomy. The taxonomy consists of fourteen fine-grained review intents, which fall into three categories according to the author's stance. These are elaborated on in Table~\ref{tab:rebuttallabels}.

\subsubsection{Rebuttal actions}

The 14 rebuttal actions (\Cref{tab:rebuttallabels}) are divided into three \textsc{rebuttal-stance} categories (\textit{concur}, \textit{dispute}, \textit{non-arg}) based on the author's stance towards the reviewer's comments.

(1) \textit{concur}: The author concurs with the premise of the context. This includes answering a question or discussing a requested change that has been made to the manuscript, conceding a criticism in an evaluative sentence.
(2) \textit{dispute}: The author disputes the premise of the context. The rebuttal sentence may reject a criticism or request, disagree with an underlying fact or assertion, or mitigate criticism (accepting a criticism while, e.g., arguing it to be offset by other properties). 
(3) \textit{non-arg}:  Encompasses rebuttal actions including \textit{social} actions (such as thanking reviewers), and \textit{structuring} labels, for sentences that organize the review.

Responses to \textit{request}s are further annotated: if the author \textit{concur}s, we record whether the task has been completed by the time of the rebuttal, or promised by the camera ready deadline; if the author \textit{dispute}s, we record whether the task was deemed to be out of scope for the manuscript.

\subsubsection{Rebuttal context}

We refer to the set of sentences which a rebuttal sentence is responding to as the \textit{context} of that sentence, with special labels for when referring to the entire review (\textit{global context}) or the empty set (\textit{no context}). By not mandating a fixed discourse chunking, these annotations may handle situations when some rebuttal sentences respond to large sections of text, and other rebuttal sentences respond to specific sentences within those sections.  

\subsection{Data Source and Annotation}

DISAPERE uses English text from scientific discussions on OpenReview~\citep{soergel2013open}, which makes peer review reports available for research purposes. We draw review-rebuttal pairs from the International Conference on Learning Representations (ICLR) in 2019 and 2020, resulting in text within the domain of machine learning research. Review-rebuttal pairs are split into train, development and test sets in a 3:1:2 ratio such that all texts associated with any manuscript occur in the same subset. Overall statistics for the dataset are summarized in \Cref{tab:data-stats}.

\input{tables/02b_data_stats}

Authors are able to respond to each ICLR review by adding a comment. Although rebuttals are not formally named, we consider direct replies by the author to the initial review comment to constitute a rebuttal.
While multi-turn interactions are possible, we focus on reviews and initial responses, and leave study of extended discussion for future work.  The text is separated into sentences using the spaCy \citep{spacy} sentence separator.

Annotation was accomplished with a custom annotation tool designed for this task, which is available as part of the code release accompanying DISAPERE. The tool is described in detail in \Cref{sec:ann_tool}. Annotators annotate each sentence of a review, then examine the rebuttal sentences in order, selecting sets of review sentences to form their context.
While this linking between sentences does not explicitly align multi-sentence chunks as in pipelined approaches to discourse alignment \citep{cheng-etal-2020-ape}, we note that since multiple sentences may be aligned to the same set of sentences in the review, some discourse structure is nevertheless latently implied. 

\subsection{Agreement}

We report Cohen's $\kappa$~\citep{cohen1960coefficient} on the IAA of labeling both review and rebuttals, treating each sentence as a labeling unit \pcref{tab:iaa}. The annotators for each example are selected randomly from the pool of 10 annotators. Cohen's $\kappa$ is calculated for sentences annotated at least twice. Where more than two annotations were produced, we calculate $\kappa$ between all pairs and normalize by the number of possible pairs. The results show between moderate and substantial chance-corrected agreement between annotators, for both \textsc{rebuttal-action} and \textsc{rebuttal-stance} labels (\Cref{app:overlap} provides details about agreement on context sentences).  While these IAA scores do illustrate the noise of the task, note that this is not highly unusual for discourse labeling tasks -- e.g. \citet{habernal2017argumentation} and \citet{miller-etal-2019-streamlined} both report  $\alpha_u$ between 0.4 and 0.5.

\begin{table}[ht]
\centering
\begin{tabular}{@{}lc@{}}
\toprule
Label                & Cohen's $\kappa$  \\ \midrule % $\gamma$
\coarse        & 0.605 \\
\fine   & 0.583   \\
\aspect              & 0.447  \\
\polarity             & 0.561 \\
\cmidrule(lr){1-2}
\rebcoarse      & 0.513  \\ % 0.547
\rebfine & 0.479  \\ \bottomrule %  0.400
\end{tabular}
\caption{\label{tab:iaa} IAA for review labels (top) and rebuttals (bottom), scored on double annotation. IAA is reported on 65 double-annotated examples, all of which fall in the test set of \myname. }
\end{table}

%% file: tables/02a_description_table.tex
\begin{table*}[t]
\resizebox{\textwidth}{!}{
\begin{tabular}{lllll}
\toprule

\multicolumn{2}{l}{Category} & Label & Description & Percentage\\ \midrule

\multicolumn{2}{l}{\multirow{6}{*}{\rotatebox{90}{\coarse}}}  & Evaluative & A subjective judgement of an aspect of the paper & 32.83\% \\
 & & Structuring & Text used to organize an argument & 27.70\% \\
 & & Request & A request for information or change in regards to the paper & 19.82\% \\
 & & Fact & An objective truth, typically used to support a claim & 8.55\% \\
 & & Social & Non-substantive text typically governed by social conventions & 1.41\% \\
 & & Other & All other sentences & 9.71\% \\ \cmidrule(lr){1-5}

\multicolumn{2}{l}{\multirow{7}{*}{\rotatebox{90}{\aspect}}}  & Substance & Are there substantial experiments and/or detailed analyses? & 17.09\% \\
 & & Clarity & Is the paper clear, well-written and well-structured? & 11.08\% \\
 & & Soundness/Correctness & Is the approach sound? Are the claims supported? & 9.58\% \\
 & & Originality & Are there new topics, technique, methodology, or insights? & 3.85\% \\
 & & Motivation/Impact & Does the paper address an important problem? & 3.69\% \\
 & & Meaningful Comparison & Are the comparisons to prior work sufficient and fair? & 3.15\% \\
 & & Replicability & Is it easy to reproduce and verify the correctness of the results? & 2.86\% \\ \cmidrule(lr){1-5}

\multicolumn{2}{l}{\multirow{2}{*}{\textsc{polarity}}}  & Negative & Negatively describes an aspect of the paper (reason to reject) & 29.43\% \\
 & & Positive & Positively describes an aspect of the paper (reason to accept) & 11.16\% \\ \cmidrule(lr){1-5}

\multirow{8}{*}{\rotatebox{90}{\textsc{\fine}}} &  \multirow{3}{*}{\rotatebox{90}{Struct.}}  & Summary & Reviewer's summary of the manuscript & 18.17\% \\
 & & Heading & Text used to organize sections of the review & 8.54\% \\
 & & Quote & A quote from the manuscript text & 1.00\% \\ \cmidrule(lr){2-5}

 & \multirow{5}{*}{\rotatebox{90}{Request}} & Explanation & Request to explain scientific choices (question) & 5.50\% \\
 & & Experiment & Request for additional experiments or results & 4.78\% \\
 & & Edit & Request to edit the text in the manuscript & 4.14\% \\
 & & Clarification & Request to clarify the meaning of some text (question) & 2.80\% \\
 & & Typo & Request to fix a typo in the manuscript & 1.98\% \\

\bottomrule
\end{tabular}}
\caption{\label{tab:reviewlabels}A list of the review sentence labels, their descriptions, and the percentage of review sentences they apply to. Labels from all categories besides \coarse are optional, and thus may not add up to 100\%.
}

\end{table*}

%% file: tables/02aa_rebuttal_description.tex
\begin{table*}
\resizebox{\textwidth}{!}{
\begin{tabular}{llllll}
\toprule

\multicolumn{2}{l}{Category} & Label & Description & Reply to & Percentage \\
 \midrule
\multirow{10}{*}{\rotatebox{90}{Argumentative}} & \multirow{6}{*}{\rotatebox{90}{Concur}}  & Answer & Answer a question & Request & 32.76\% \\
& & Task has been done & Claim that a requested task has been completed & Request & 8.58\% \\
& & Concede criticism & Concede the validity of a negative eval. statement & Evaluative & 2.70\% \\
& & Task will be done & Promise a change by camera ready deadline & Request & 2.01\% \\
& & Accept for future work & Express approval for a suggestion, but for future work & Request & 1.30\% \\
& & Accept praise & Thank reviewer for positive statements  & Evaluative & 0.35\% \\ \cmidrule(lr){2-6}
& \multirow{4}{*}{\rotatebox{90}{Dispute}}& Reject criticism & Reject the validity of a negative eval. statement & Evaluative & 10.37\% \\
& & Mitigate criticism & Mitigate the importance of a negative eval. statement & Evaluative & 2.43\% \\
& & Reject request & Reject a request from a reviewer & Request & 1.16\% \\
& & Refute question & Reject the validity of a question & Request & 0.95\% \\
& & Contradict assertion & Contradict a statement presented as a fact & Fact & 0.86\% \\ \midrule
 \multirow{4}{*}{\rotatebox{90}{Non-arg}}& & Structuring & Text used to organize sections of the review & - & 17.82\% \\
& & Summary & Summary of the rebuttal text & - & 7.94\% \\
& & Social & Non-substantive social text & - & 6.71\% \\
& & Followup question & Clarification question addressed to the reviewer & - & 0.32\% \\
& & Other & All other sentences & - & 3.75\% \\

\bottomrule
\end{tabular}
}
\caption{
\label{tab:rebuttallabels}A list of the rebuttal sentence labels, their descriptions, and the percentage of rebuttal sentences they apply to. The ``Reply to'' column shows the \coarse types that a particular rebuttal type would canonically reply to. Each rebuttal sentence has exactly one \rebfine label, so these percentages add up to 100\%.}
\end{table*}

%% file: tables/02b_data_stats.tex
\begin{table}[ht]

\resizebox{\columnwidth}{!}{
\begin{tabular}{llll}
\toprule
{} &  Train &    Dev &   Test \\
\midrule
Num. review-rebuttal pairs &    251 &     88 &    167 \\
Num. manuscripts             &     94 &     37 &     57 \\
Num. adjudicated pairs  &      0 &      0 &     65 \\
Num. review sentences   &   5216 &   1484 &   3246 \\
Num. rebuttal sentences &   5805 &   2015 &   3283 \\
%Avg. sents per review   &  19.31 &  15.43 &  18.49 \\
%Avg. sents per rebuttal &  23.13 &   22.9 &  19.66 \\
Num. review tokens   &   112k &   33k &   70k \\
Num. rebuttal tokens &   131k &   44k &   75k \\
\bottomrule
\end{tabular}}
 \caption{Statistics for the dataset. Where possible, multiple reviews for the same manuscript were annotated. All reviews for any particular manuscript fall within the same train/dev/test split. Adjudicated pairs are those that were annotated by multiple annotators, and had disagreements resolved by an experienced annotator. All test set pairs are double-annotated. While the original sentence boundaries were maintained, tokenization within sentences was carried out using Stanza\cite{qi2020stanza}.}
\label{tab:data-stats}
 \end{table}

%% file: 04_analysis.tex
\section{Analysis}
\label{sec:analysis}

\subsection{Context types}

We separate the different types of rebuttal contexts in terms of the number and relative position of selected review sentences in \Cref{tab:context-types}, along with the four cases in which  the context cannot be described as a subset of review sentences. Notably, 84.81\% of sentences are linked to some review context. A small number of sentences refer to other sentences within the rebuttal, rather than any review context, posing a challenge for future work.

\begin{table}[ht]
    \centering
    \resizebox{\columnwidth}{!}{
    \begin{tabular}{llrl}
\toprule
 &\multirow{2}{*}{Context type} & \multicolumn{2}{l}{Rebuttal sents.} \\
                             & & (Num.)     & (\%)        \\
\midrule
\multirow{3}{*}{\rotatebox{90}{\makecell{Sents. \\ selected}}} &   Multiple contiguous &           4696 &      42.29\% \\
    &    Single sentence &            4313 &      38.85\% \\
&Mult. non-contiguous &            407 &      3.67\% \\ \cmidrule(lr){1-4}
\multirow{3}{*}{\rotatebox{90}{\makecell[cm{2.4cm}]{\centering No sents. \\ selected}}} &       Global context &             816 &       7.35\% \\
   & Context in rebuttal &             647 &       5.83\% \\
    &         No context &             152 &       1.37\% \\
     &     Context error &             61 &       0.55\% \\
&  Cannot be determined &              11 &       0.10\% \\
\bottomrule
\end{tabular}}
    \caption{Different types of rebuttal sentence contexts. Top: Over 84\% of sentences are linked to a subset of sentences in the review. Bottom: sentences not linked to any particular subset of review sentences.}
    \label{tab:context-types}
\end{table}

\subsection{Alignment}

\begin{figure}[ht]
    \centering
    \includegraphics[width=\columnwidth]{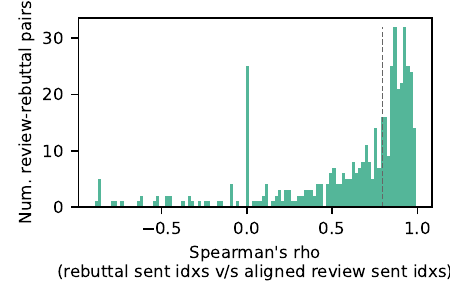}
    \caption{Spearman's $\rho$ between rebuttal sentence indices and aligned review sentence indices. The dashed line indicates the median $\rho$ value, which falls at $0.794$.}
    \label{fig:spearmen}
\end{figure}

One might reasonably hypothesize that the task of alignment between rebuttal and review sentences would be trivial, since authors are likely to respond to each point in the review in order. We can show that this is not the case. In \Cref{fig:spearmen}, we calculate Spearman's $\rho$ between rebuttal sentence indices and their aligned review sentence indices. Rebuttals responding to each point in order would achieve $\rho = 1.0$; this case is rare. Many examples with positive $\rho < 1.0$ indicate that authors do respond to points approximately in order, but a simple mapping based on order alone would not capture the correct alignment. Thus, while linear inductive bias may be beneficial to alignment models, the task of determining rebuttal sentences' contexts is not trivial.

\subsection{Author interpretations of criticism}

\begin{figure}[t]
    \centering
    \includegraphics[width=\columnwidth]{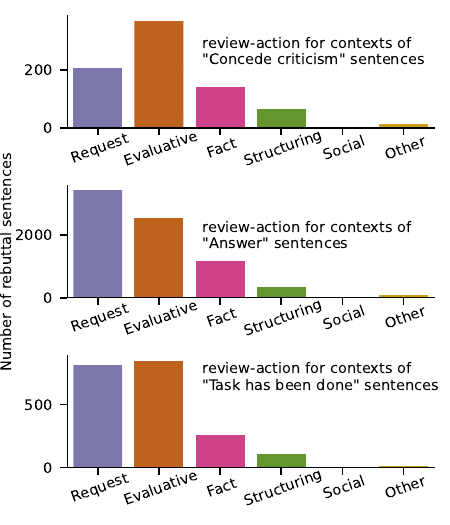}
    \caption{Distribution over \coarse for the context sentences of three \textsc{rebuttal-action}s. The canonical \coarse is marked by cross hatching. Note that authors sometimes interpret requests as criticisms (``Concede criticism''); often respond to evaluative sentences as if they are questions (``Answer''), and sometimes treat criticisms in the form of evaluative sentences as requests which they then carry out. (``Task has been done'')}
    \label{fig:noncanonical}
\end{figure}

In our taxonomy, each argumentative \textsc{rebuttal-action} corresponds to a particular \textsc{review-action}, which we refer to as its \textit{canonical} \textsc{review-action} (listed in `Reply to' column of \Cref{tab:rebuttallabels}). For example, \textit{answers} are generally responses to \textit{requests}, while \textit{conceding criticism} is usually a response to an \textit{evaluative} statement. Annotations revealed that authors often interpreted review sentences as if they embodied \textsc{review-action}s besides the canonical one, in a way that furthered the author's argumentative goal. For example, authors often responded to \textit{evaluative} statements as if they were \textit{requests}, perhaps in order to appease a reviewer, although no action was explicitly requested. \Cref{fig:noncanonical} shows the distribution of contexts for three different \textsc{rebuttal-action}s.

\subsection{Relating discourse features to rating}
\begin{figure}[t]
    \centering
    \includegraphics[width=\columnwidth]{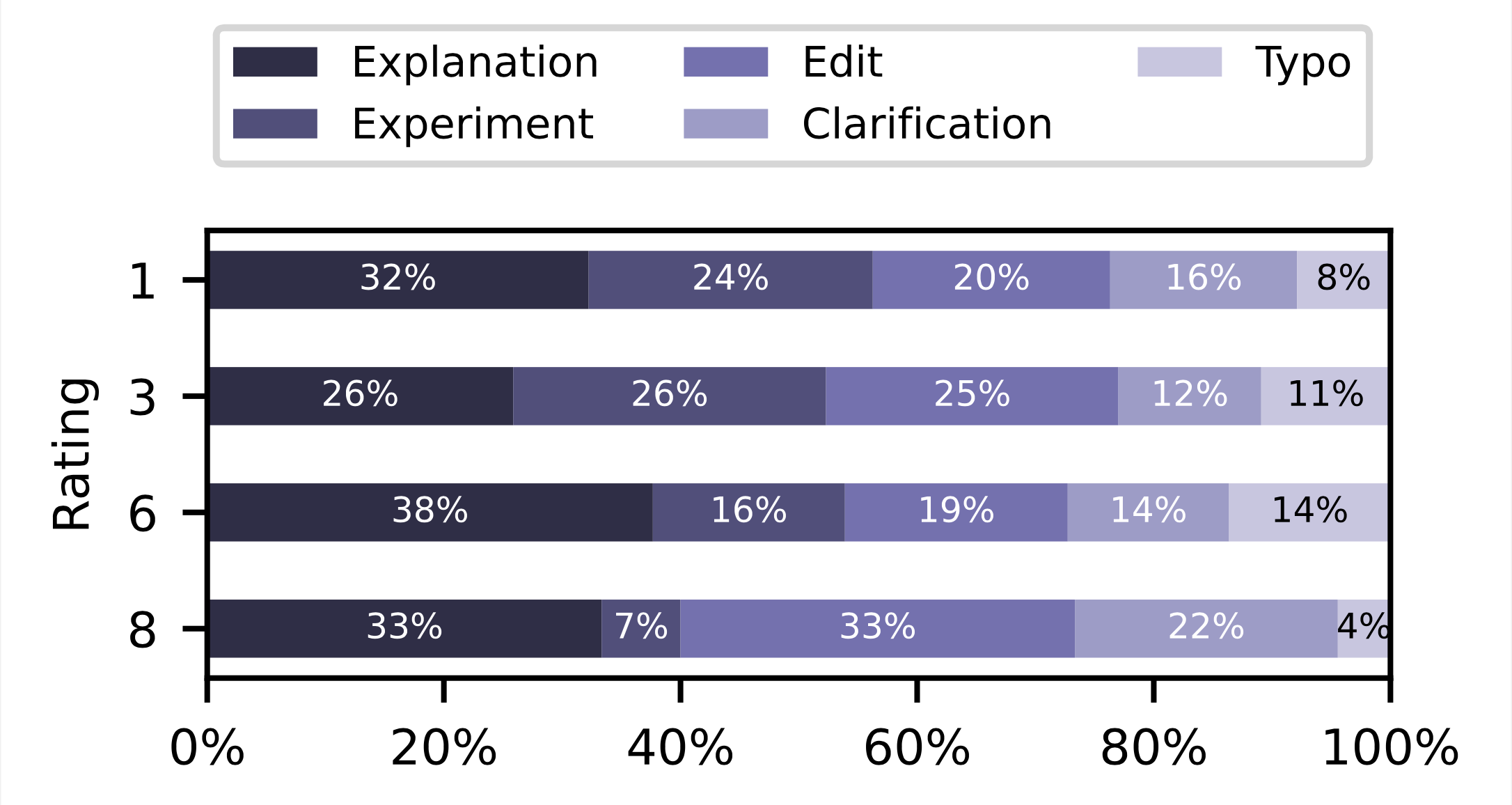}
    \caption{Distribution of \coarse labels, separated by rating}
    \label{fig:action-rating}
\end{figure}

\Cref{fig:action-rating} shows one possible analysis taking into account the rating of the review. We show the distribution of \fine labels of \textit{requests} with review ratings. It appears that high-scoring manuscripts are rarely asked to add experiments, and are polished enough to not elicit requests to fix typos. Interestingly, low-scoring manuscripts have the second-lowest occurrence of typo requests, which could be due to the preponderance of other requests, but this bears further examination.

% \subsection{Reply coverage}
% Fig.~\ref{fig:coverage} shows the distribution of reply coverage for reviews, i.e. the percentage of review sentences that received a direct reply in the rebuttal. While the overall trend is unremarkable, considering different review actions separately reveals a bimodal distribution for both request and fact statements, indicating that authors either exhaustively reply to these statements in a point-by-point fashion, or ignore or accept them implicitly.

% \begin{figure}[h]
%     \centering
%     \includegraphics[width=\columnwidth]{figs/coverage.pdf}
%     \caption{Histograms over all reviews based on the the proportion of sentences receiving direct replies, as annotated in the rebuttal. Three of the histograms show coverage taking into account only request, evaluative and fact sentences respectively.}
%     \label{fig:coverage}
% \end{figure}

%% file: 05_application.tex
\section{Application: Agreeability}
\label{sec:applications}

\citet{gao-etal-2019-rebuttal} showed that reviewers do not appear to act upon the rebuttals responding their reviews. It is possible that this is due to paucity of time on the reviewers' part. It is also common practice for area chairs to use review variance across a manuscript's reviews as a practical heuristic to decide which manuscripts need their attention. We propose that discourse information such as that described by \myname\ can be used to provide heuristics that are data-driven, yet interpretable, and leverage information from the content of reviews rather than just numerical scores, resulting in better decision making.

\begin{figure}[ht]
    \centering
    \includegraphics[width=\columnwidth]{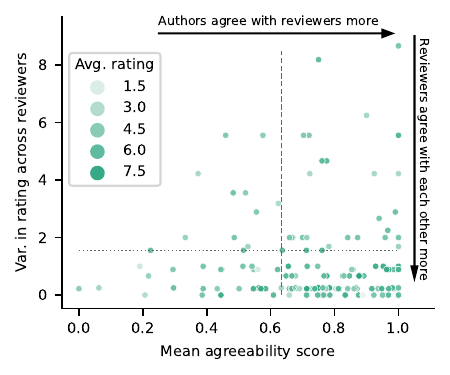}
    \caption{Mean agreeability for a manuscript's reviews v/s reviewer variance. Manuscripts above the dotted line are in the top quartile of rating variance, and are more likely to be reviewed by area chairs. Manuscripts to the left of the dashed line are in the bottom quartile of mean agreeability, in which authors take issue with the premises of reviewers' comments. The color of the dots indicates the mean of the reviewers' ratings.}
    \label{fig:agreeability}
\end{figure}

One such measure is \textit{agreeability}, which we define as the ratio of \textsc{concur} sentences to argumentative sentences in a rebuttal, i.e.: $\mathit{agreeability} = \frac{n_{\mathit{concur}}}{n_{\mathit{concur}} + n_{\mathit{dispute}}}$. We argue that low agreeability can indicate problematic reviews even in cases where the variance in scores does not reveal an issue, as illustrated in \Cref{fig:agreeability}.  Agreeability is only weakly correlated with rating, with Pearson's $r=0.347$. In \Cref{fig:agreeability}, 18\% (28/159) of manuscripts would not meet the bar for high variance scores (top quartile), although their low agreeability (bottom quartile) indicates that they may merit closer attention from area chairs\footnote{Two such examples included in DISAPERE: \url{https://openreview.net/forum?id=r1e74a4twH} and \url{https://openreview.net/forum?id=HyMRUiC9YX}.}.

%% file: 06_experiments.tex
\section{Baselines}
\label{sec:baselines}

Two types of machine learning tasks can be defined in DISAPERE. First, a sentence-level classification task for each of the four review labels and the two levels of rebuttal labels. Second, an alignment task in which, given a rebuttal sentence, the set of review sentences that form its context are to be predicted.

The models described below are not intended to introduce innovations in discourse modeling, rather, we intend to show the off-the-shelf performance of state-of-the-art models, and indicate through error analysis the phenomena that are yet to be captured.

\subsection{Sentence classification}

For the six classification tasks, we use \texttt{bert-base} \citep{devlin-etal-2019-bert} to produce sentence embeddings for each sentence, then classify the representation of the \texttt{[CLS]} token using a feedforward network.

We report macro-averaged F1 scores, shown in \Cref{tab:class-results}. In general, F1 is lower for tasks with larger label spaces. While the performance is reasonable in most cases, there is still room for improvement. While \aspect achieves a particularly low F1 score, its $\kappa$ is within the bounds of moderate agreement; thus, this must be accounted for by the inherent difficulty of the task rather than a deficit in data quality.

\begin{table}[htbp]
\centering
\small
% \resizebox{\columnwidth}{!}{}
\begin{tabular}{@{}l l c c@{}}
\toprule
Classification task   	& \makecell{Macro F1\\ (test)} & \makecell{Cohen's \\$\kappa$} & \makecell{Num.\\ labels} \\ \midrule
\coarse 	& 60.42\% & 0.605                 & 7           \\
\fine     	& 44.83\% & 0.583                 & 10          \\
\aspect    	& 38.28\% & 0.447                 & 9           \\
\polarity   & 70.88\% & 0.561                 & 3           \\ \cmidrule(lr){1-4}
\rebcoarse  & 43.36\% & 0.513                 & 4           \\
\rebfine   	& 31.23\% & 0.479                 & 17          \\ \bottomrule
\end{tabular}
\caption{Sentence classification results. Top: review labels; Bottom: rebuttal labels. }
\label{tab:class-results}
\end{table}

\begin{table*}[htbp]
\small
\begin{tabular}{llc}\toprule
& & Label (Pred.) \\ \midrule
1 & Can the proposed [...] function represent all function the authors used in the paper? \textit{Yes.}     &  E (R)                                                   \\
2 & Matrices can have either “horizontal” or “vertical” redundancy (or “other” or neither).            & E (F)                    \\ 
3 & Solid technical innovation/contribution:                                                        & E                        \\
4 & I am also wondering if the comparison with the baselines is fair.                           & E                            \\ 
5 & I wonder if the authors ever looked at how much [...] determines the performance of the system? & R \\\bottomrule
\end{tabular}
\caption{Example sentences including errors and challenging cases. E, R, F stand for \textit{evaluative}, \textit{request} and \textit{fact} respectively. Letters in parentheses show the incorrect label from the model. Sentence (3) functions both as \textit{evaluative} and \textit{structuring}. Sentences (4) and (5) share a prefix but have different \textsc{review-action}s.}
\label{tab:errors}
\vspace{-4mm}
\end{table*}

As one might expect, errors in the classification results largely mirror disagreements in the annotations, which in turn reflect particularly ambiguous utterances. One example is the occurrence of rhetorical questions, such as (1) in \Cref{tab:errors}, incorrectly labeled as \textit{request} instead of \textit{evaluative}. In fact, for sentences such as (1), additional context would disambiguate its type: the reviewer answers the question in the next sentence, and hence both sentences were labeled \textit{evaluative}. Similarly, (2) was labeled \textit{fact}, but since it is an integral part of a reviewer’s argument against the soundness of the paper, should have been labeled \textit{evaluative}. 
Certain reviewers also use conventions that do not fit the general schema we observed when developing DISAPERE. For example, (3), an opinionated heading, could be considered both \textit{structuring} and \textit{evaluative}.
Finally, certain lexical cues a model may pick up on can be quite subtle. For example, though they share a prefix, sentences (4) and (5) are clearly \textit{evaluative} and \textit{request} respectively.

\subsection{Rebuttal context alignment}

We model rebuttal context alignment as a ranking task. Ideally, a model should rank all relevant review sentences higher than non-relevant review sentences. As a baseline, we use an information retrieval (IR) model based on BM25 that, given a rebuttal sentence ranks all the corresponding review sentences. We also report results from a neural sentence alignment model based on a two-tower Siamese-BERT (S-BERT) model \cite{reimers-2019-sentence-bert}. We add a \texttt{NO\_MATCH} sentence to the review, to which rebuttal sentences without context sets in the review are aligned. Then, each review and rebuttal sentence is encoded independently using a S-BERT encoder and the similarity between two sentences is computed using cosine similarity. We initialize with a model\footnote{We initialize from a \hyperlink{https://huggingface.co/sentence-transformers/all-MiniLM-L6-v2}{sentence-transformers/all-MiniLM-L6-v2} model} pre-trained on various sentence-pair datasets. Alignment is evaluated using mean reciprocal rank (MRR) and Mean Average Precision (MAP). 

\begin{table}[ht]
\centering
% \small
\begin{tabular}{@{}lll@{}}
\toprule
    & S-BERT & BM25  \\ \midrule
MAP & 0.4409 & 0.5174 \\
MRR & 0.5022 & 0.5980 \\ \bottomrule
\end{tabular}
\caption{Rebuttal context alignment results. The results of both models indicate significant scope for improvement.}
\label{tab:RR}
\vspace{-4mm}
\end{table}

Surprisingly, the BM25 model outperforms a neural model \citep{beir}. While this shows that lexical information is a useful signal, both models have significant scope for improvement, and lexical overlap is clearly not sufficient for this task. Importantly, neither of these models account for the context of the rebuttal sentence, and predict each sentence's context independently. Incorporating this information is likely to lead to performance gains; however, we leave this investigation to future work.

%% file: 07_conclusion.tex
\section{Conclusion}
As the burden of academic peer reviewing grows, it is important for program chairs and editors to act upon data-driven insights rather than heuristics, to make the best possible use of participants' scarce time. Models trained on data like DISAPERE will allow decision makers to glean deep insights on the interactions occurring during peer review.

Almost all publicly available peer review data is from the domain of artificial intelligence, limiting the scope of DISAPERE and any similar project. While this means that models trained on DISAPERE won't necessarily generalize to all new domains, we hope that with the detailed annotation guidelines and seamless data collection using the software provided with this paper support, users can build on our work, and ensure that their insights are robust to differences over time and across fields.

\section{Ethics}

The outcomes of peer review can have outsize effects on the careers of participating scholars. As machine learning models are known to amplify biases, we strongly recommend against using the outputs of any machine learning system to make decisions about individual cases. A dataset like DISAPERE is best used to survey participants' behavior. Any interventions based on this information should be subjected to studies in order to ensure that they do not introduce or exacerbate bias.

%% file: 10_acknowledgements.tex
\section*{Acknowledgments}

This material is based upon work supported in part by the National Science Foundation under Grant Numbers IIS-1763618, IIS-1922090, and IIS-1955567, in part by the Defense Advanced Research Projects Agency (DARPA) via Contract No. FA8750-17-C-0106 under Subaward No. 89341790 from the University of Southern California, in part by the Office of Naval Research (ONR) via Contract No. N660011924032 under Subaward No. 123875727 from the University of Southern California, in part by IBM Research AI through the AI Horizons Network, in part by the Chan Zuckerberg Initiative under the project Scientific Knowledge Base Construction, and in part by the Center for Intelligent Information Retrieval. Any opinions, findings and conclusions or recommendations expressed in this material are those of the authors and do not necessarily reflect those of the sponsors.

% \begin{itemize}
%     \item the National Science Foundation under Grant Numbers IIS-1763618, IIS-1922090, IIS-1955567
%     \item the Defense Advanced Research Projects Agency (DARPA) via Contract No. FA8750-17-C-0106 under Subaward No. 89341790 from the University of Southern California 
%     \item the Office of Naval Research (ONR) via Contract No. N660011924032 under Subaward No. 123875727 from the University of Southern California
%     \item IBM Research AI through the AI Horizons Network
%     \item the Chan Zuckerberg Initiative under the project Scientific Knowledge Base Construction
    
% \end{itemize}

%% file: 08_appendices.tex
\newpage

\section{Rationale for taxonomy construction}
\label{sec:rationale}
Our label sets leverage ideas from and commonalities between existing work in this domain, including AMPERE \citep{hua-etal-2019-argument-mining}, AMSR \citep{fromm2020argument} ASAP-Review \citep{yuan2021automate}, and \citet{gao-etal-2019-rebuttal}:
\begin{itemize}
    \item ASAP-Review's polarity labels approximately correspond to \textit{arg-pos} and \textit{arg-neg} labels in AMSR
    \item AMSR and AMPERE each label non-argumentative sentences in a similar manner
    \item \textit{aspect} labels from ASAP-Review apply only to certain types of sentences; namely \textit{request} and \textit{evaluative} sentences from AMPERE's taxonomy.
    \item \textit{summary} is an exception among ASAP-Review's \textit{aspect}s, behaving similarly to AMPERE's \textit{quote}. We thus include both of these under a \textit{structuring} category.
    \item Further, in order to gauge the extent to which authors acquiesced to reviewers' requests, we introduce a fine-grained categorization of the types of requests.
    \item \citet{gao-etal-2019-rebuttal} enumerates some features of rebuttals, including expressing gratitude, promising revisions, and disagreeing with criticisms. We formalize these observations into our rebuttal label taxonomy.
\end{itemize}

\section{Annotation tool}
\label{sec:ann_tool}

Two modes of annotation are possible. First, annotators can apply labels on a sentence-by-sentence basis. Multiple labeling schemas can be annotated simulatenously, with the option of adding constraints so that certain values govern possible values for other properties.  This annotation mode is shown in \Cref{fig:review_annot}.

The second annotation mode can build on the output of the first annotation mode. Here, sentences of a focus text (the rebuttal) are presented in sequence, and annotators are permitted to select one or more of the sentences in the reference text (the review) which form the context of the sentence of the focus text. Further, a label can be applied to the alignment. This annotation mode is shown in \Cref{fig:rebuttal_annot1} and \Cref{fig:rebuttal_annot2}.

\begin{center}
\begin{figure*}
  \includegraphics[width=\textwidth]{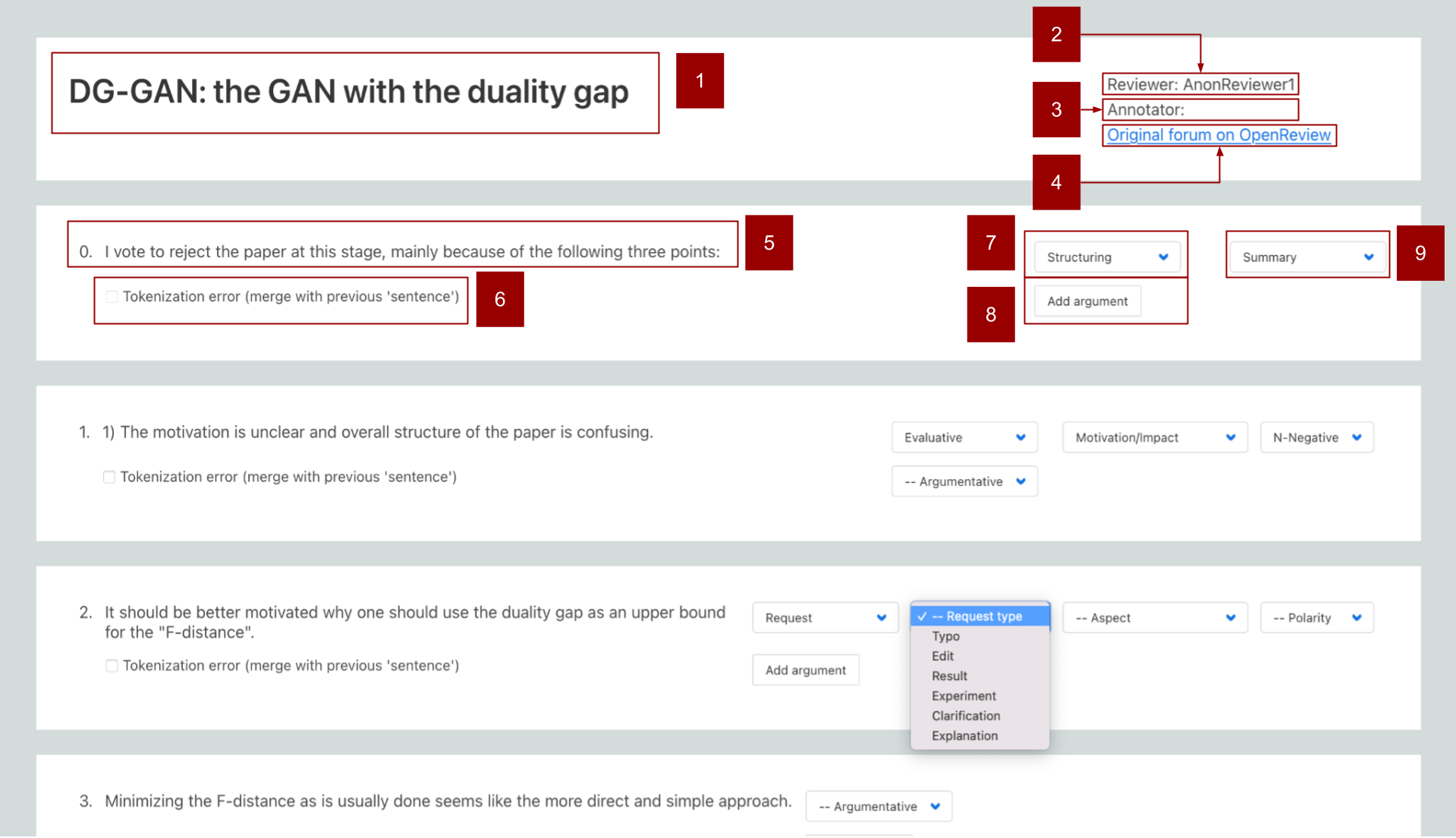}
  \caption{\label{fig:review_annot}Review annotation interface. Annotators select label values from dropdown menus for each review sentence in turn.
[1] Title of the manuscript whose review is being annotated
[2] Reviewer identifier
[3] Annotator identifier (removed for anonymity)
[4] Link to original forum, in case it is needed for context
[5] Individual review sentence
[6] Option to report sentence splitting error (sentence splitting generally suffered from false positives)
[7] Dropdown for \textsc{review-action}
[8] Follow-up dropdownfor \textsc{fine-review-action} populated based on value in (7)
[9] Button to add second \textsc{review-action} if necessary (this was seldom used)}
\end{figure*}

\begin{figure*}
  \includegraphics[width=\textwidth]{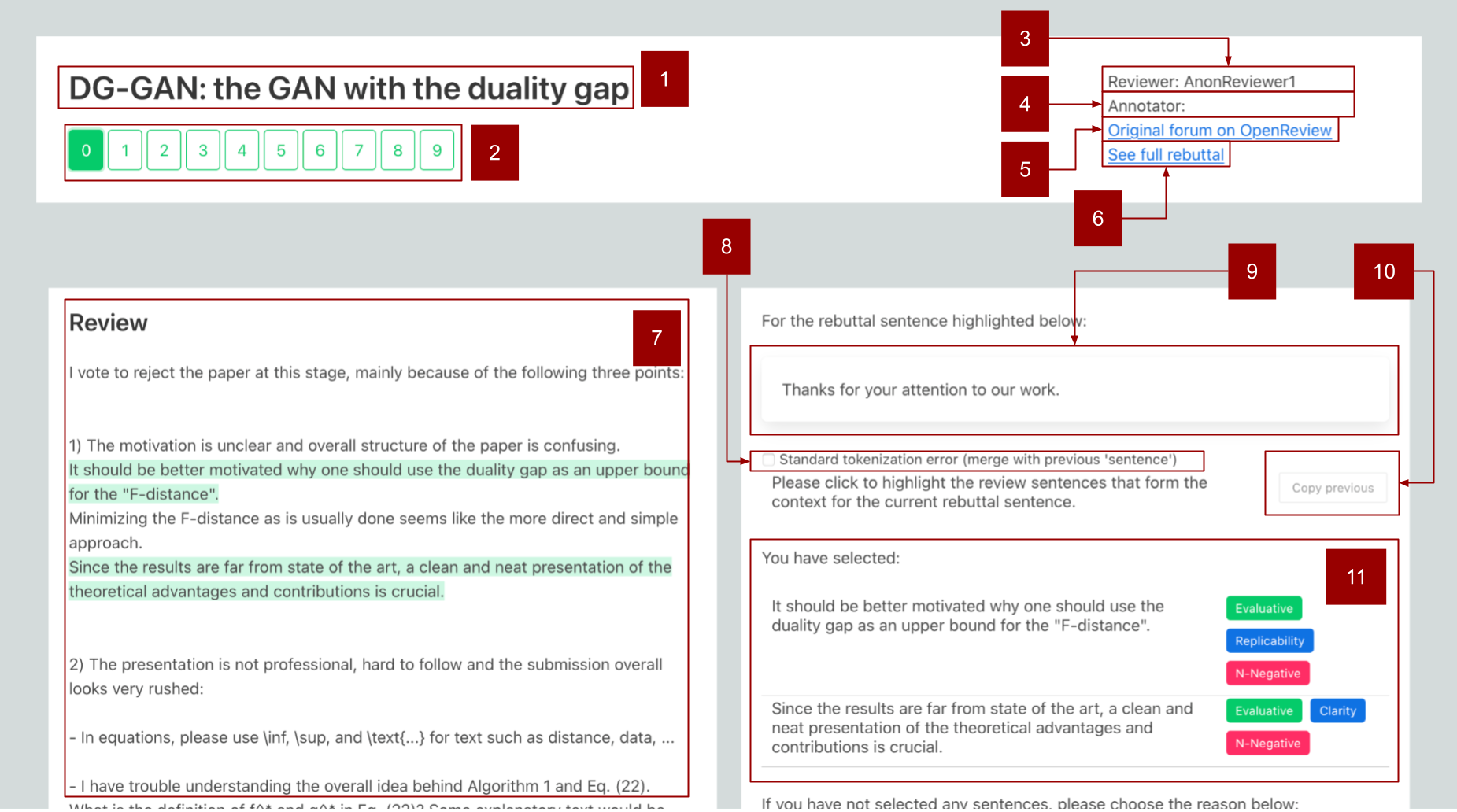}
  \caption{\label{fig:rebuttal_annot1}Rebuttal annotation interface. Annotators examine each rebuttal sentence in turn, selecting sentences as context and specifying \textsc{rebuttal-action}.
[1] Title of the manuscript whose review is being annotated
[2] Buttons to navigate between rebuttal sentences. Each page refers to a single rebuttal sentence (See (9))
[3] Reviewer identifier
[4] Annotator identifier (removed for anonymity)
[5] Link to original forum, in case it is needed for context
[6] Link to open pop-up window with full rebuttal text, in case it is needed for context
[7] Full review text. When a review sentence is clicked, it is highlighted and its details appear in (11)
[8] Option to report sentence splitting error (false positive)
[9] Rebuttal sentence being annotated
[10] Button to copy \textsc{rebuttal-action} label and context from previous rebuttal sentence
[11] Table showing details of selected context sentences from the review, with the labels the annotator provided
The screenshot is continued in \Cref{fig:rebuttal_annot2}.}
\end{figure*}

\begin{figure*}
  \includegraphics[width=\textwidth]{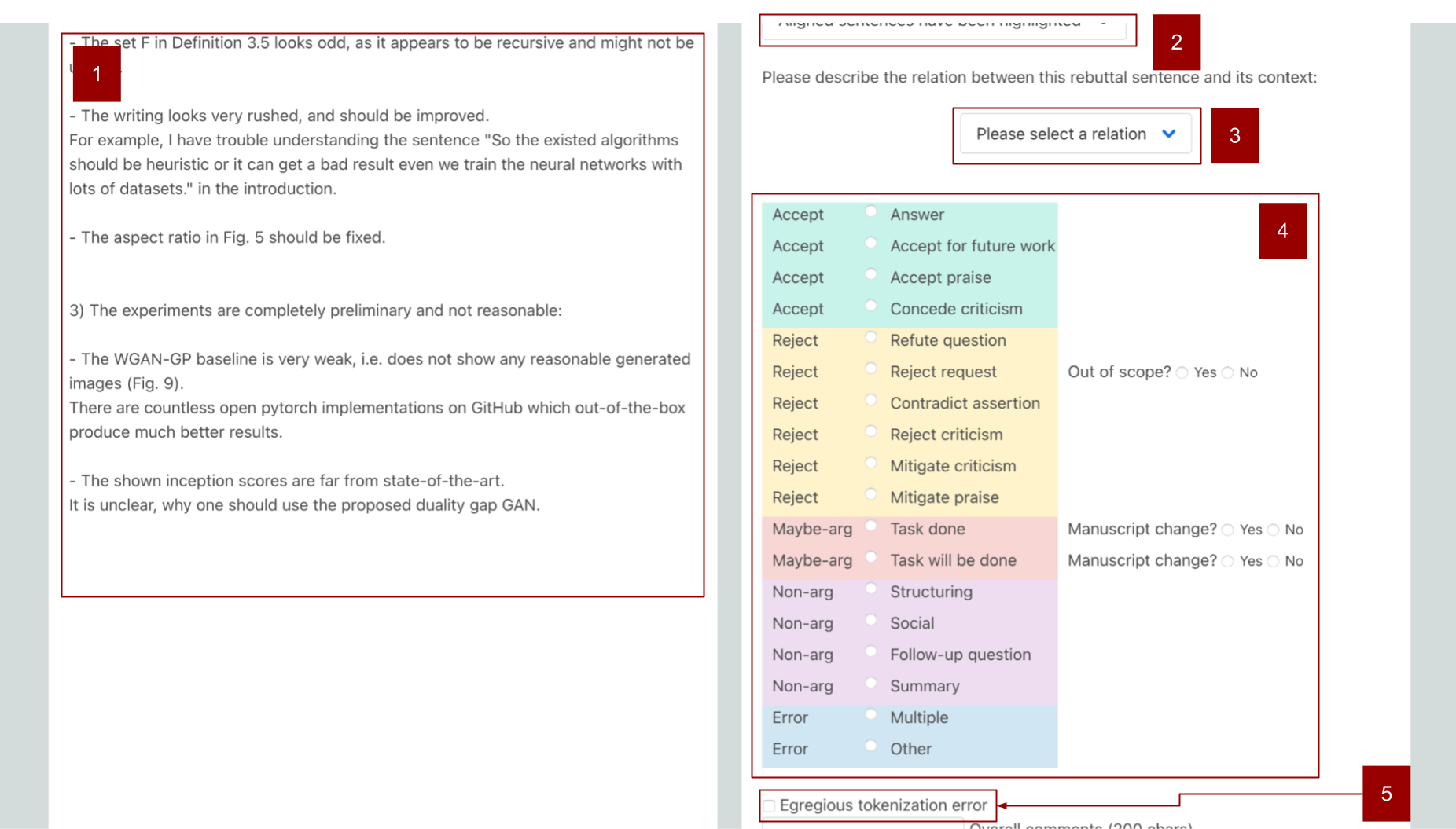}
  \caption{\label{fig:rebuttal_annot2}Rebuttal annotation interface (continued from \Cref{fig:rebuttal_annot1}). Annotators examine each rebuttal sentence in turn, selecting sentences as context specifying \textsc{rebuttal-action}.
[1] Full review text, continued from (7) in \Cref{fig:rebuttal_annot1}.
[2] Dropdown to select context type, in case context cannot be defined as a subset of review sentences.
[3] Dropdown to select \textsc{rebuttal-action} (keyboard navigation possible)
[4] Buttons to select \textsc{rebuttal-action} (in case mouse navigation is preferred)
[5] Option to report egregious sentence splitting errors.}
\end{figure*}

\end{center}

\section{Annotated review-rebuttal pair}

\Cref{app:example} shows a truncated version of a review-rebuttal pair from the train set of DISAPERE.

\input{app_example}

\section{Context overlap analysis}
\label{app:overlap}

As a proxy for agreement of rebuttal spans, we show the types of overlap between spans on rebuttal sentences from 81 examples annotated by two annotators in \Cref{tab:context-overlap}.

\begin{table}[h]
\begin{tabular}{@{}lll@{}}
\toprule
\makecell{Type of\\ context overlap} & \makecell{Num. rebuttal\\ sentences} & \makecell{\% rebuttal \\sentences} \\ \midrule
Exact match             & 914                    & 53.11\%               \\
Partial match           & 492                    & 28.59\%               \\
Agree none              & 122                     & 7.09\%               \\
Disagree none           & 100                     & 5.81\%                \\
No overlap              & 93                     & 5.40\%                \\ \bottomrule
\end{tabular}
\caption{\label{tab:context-overlap} Types of context overlap. Full agreement is achieved in the top rows (exact match and `Agree none', where both annotators agree that there is no appropriate subset of review sentences forming the context. in `Disagree none', one annotator marks a subset of review sentences, while the other does not.}
\end{table}

\section{Additional Agreement Analysis}
\begin{figure*}
  \includegraphics[width=\textwidth]{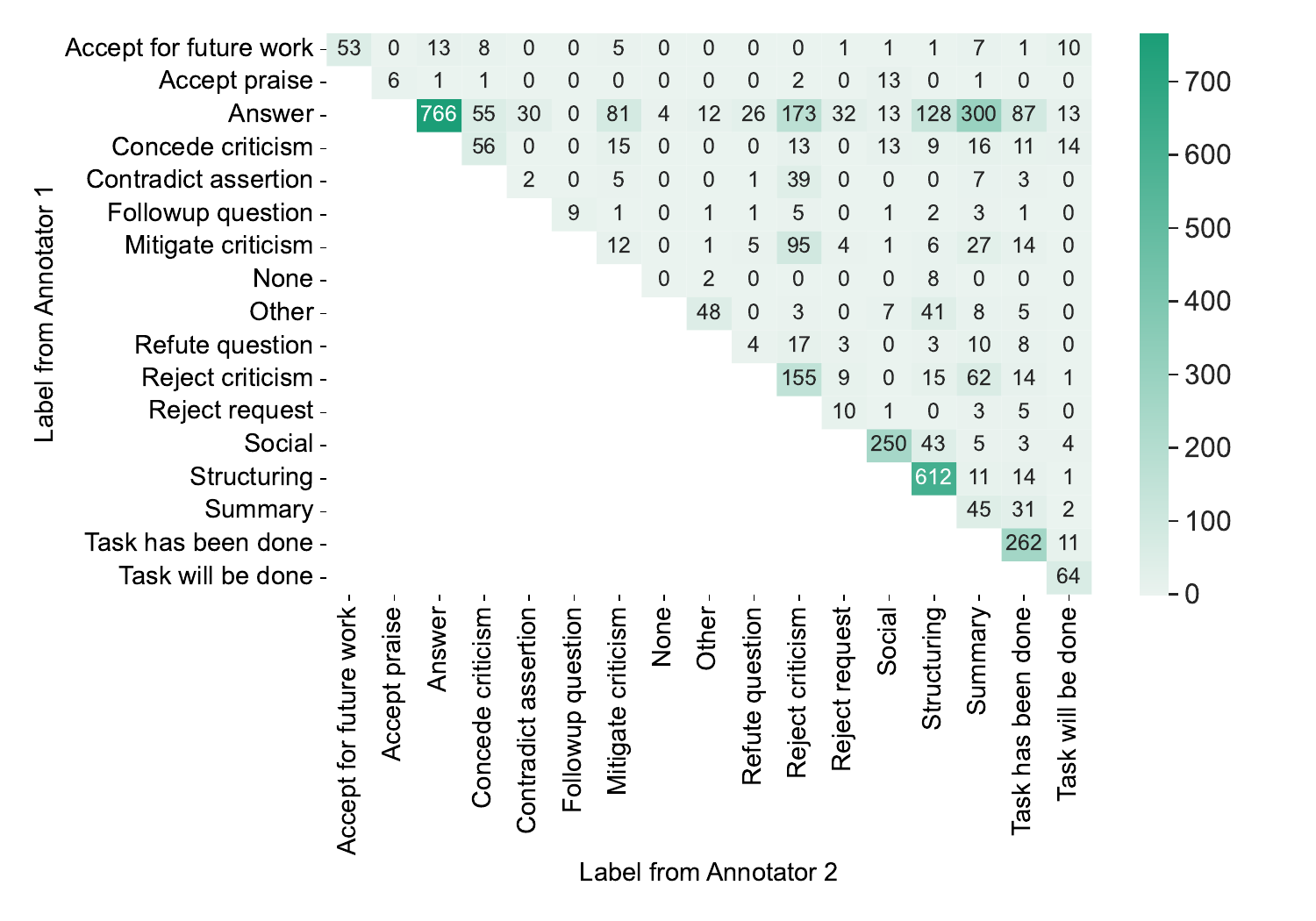}
  \caption{\label{fig:reb_fine_confusion} Confusion matrix showing agreement between annotators for \textsc{rebuttal-action} labels. }
\end{figure*}

While some of the IAA scores on annotation are low, we note that the labels used in this task attempt to characterize relatively complex relationships in text.  To give more insight into such disagreements, \Cref{fig:reb_fine_confusion} provides a confusion matrix regarding the  \rebfine labels.
Recognizing that there are often situations in which users of a dataset will hope to reduce a label set, we provide some guidance as to which such merges may be acceptable and which are not.

Many disagreements come from three labels which might be said to exist upon a continuum -- \textsc{answer}, \textsc{mitigate criticism} and \textsc{reject criticism}.  We suggest that in the situation of needing to minimize IAA disagreement, one might consider first merging \textit{mitigate criticism} into \textit{reject criticism}. The kind of disagreements seen between the two are understandable but nuanced: the difference between saying that the reviewer has a point (but that they disagree on the relevance of that point) and disagreeing with the point itself.  Out-of-context rebuttal sentences illustrating this are provided below as examples of this kind of ambiguous situation:

\begin{itemize}
\item  \textit{We note that such rules are indeed limited to some extent, but they still capture a rather expressive fragment of answer set programs with restricted forms of external computations.}
\item \textit{The use of $C_{p}^{val}$ for hyperparameter tuning was incidental and not a central point of our paper.}
\item \textit{We agree that the measure theoretic approach is not always necessary (indeed for angular actions, it is not needed), but it is necessary for a very common scenario -- clipped actions.}
\end{itemize}

Furthermore, we note that (as illustrated in the confusion matrix) a wide range of disagreements are hard to distinguish from ``answer'' labels, as authors often attempt to frame disagreements as simple answers to questions. 

%% file: app_example.tex
\begin{figure*}
\begin{framed}
  \begin{minipage}{\textwidth}
  {\Large
\begin{lstlisting}
{
  "metadata": {
    "forum_id": "ryGWhJBtDB",
    "review_id": "BJgmhEfTcH",
    "rebuttal_id": "rye3zaZ7or",
"title": "Hyperparameter Tuning and Implicit Regularization in Minibatch SGD",
    "reviewer": "AnonReviewer3", "rating": 3, "conference": "ICLR2020",
    "permalink": "https://openreview.net/forum?id=ryGWhJBtDB&noteId=rye3zaZ7or",
    "annotator": "anno10"
  },
  "review_sentences": [
    {
      "review_id": "BJgmhEfTcH",
      "sentence_index": 0,
      "text": "This paper is an empirical contribution regarding SGD arguing that it presents two different behaviors which the authors name a noise dominated regimen, and a curvature dominated regime.",
      "suffix": "",
      "review_action": "arg_structuring", "fine_review_action": "arg-structuring_summary",
      "aspect": "none", "polarity": "none"
    },
...
    {
      "review_id": "BJgmhEfTcH",
      "sentence_index": 4,
      "text": "I find the observations interesting, but the contribution is empirical and not entirely new. It would be nice if there were some theoretical results to back up the observations.",
      "suffix": "",
      "review_action": "arg_evaluative", "fine_review_action": "none",
      "aspect": "asp_originality", "polarity": "pol_negative"
    }
  ],
  "rebuttal_sentences": [
    {
      "review_id": "BJgmhEfTcH", "rebuttal_id": "rye3zaZ7or",
      "sentence_index": 0,
      "text": "We thank the reviewer for their comments.",
      "suffix": "\n\n",
      "rebuttal_stance": "nonarg", "rebuttal_action": "rebuttal_social",
      "alignment": [ "context_global", null]
    },
    {
      "review_id": "BJgmhEfTcH", "rebuttal_id": "rye3zaZ7or",
      "sentence_index": 1,
      "text": "Although our primary contributions are empirical, we also provided a detailed theoretical discussion in section 2, where we give a clear and simple account of why the two regimes arise.",
      "suffix": "",
      "rebuttal_stance": "dispute", "rebuttal_action": "rebuttal_reject-criticism",
      "alignment": ["context_sentences", [4]]
    },
    ...
  ]
}
\end{lstlisting}}
  \end{minipage} \quad
  \end{framed}
\caption{\label{app:example}A (truncated) example from the training set of DISAPERE.}
\end{figure*}